\documentclass{article}

\usepackage{caption}
\usepackage{subcaption}

\usepackage{booktabs}

\usepackage{amsmath}
\usepackage{amsfonts}
\usepackage{amssymb}
\usepackage{amsthm}

\usepackage{amsmath, amsthm, amssymb, bm}
\usepackage{dsfont}
\usepackage{amsfonts}       
\usepackage{nicefrac}       
\usepackage{microtype}      

\newcommand{\x}{\mathbf{x}}
\newcommand{\f}{\mathbf{f}}
\newcommand{\varration}{\text{variation-ratio}}

\newcommand{\bbH}{{\mathbb H}}
\newcommand{\D}{\mathcal{D}}
\newcommand{\train}{{\text{train}}}
\newcommand{\bbI}{{\mathbb I}}
\newcommand{\bo}{\text{\boldmath$\omega$}}
\newcommand{\E}{{\mathbb E}}
\newcommand{\ph}{\widehat{p}}
\newcommand{\bph}{\widehat{\mathbf{p}}}
\newcommand{\boh}{\widehat{\bo}}
\newcommand{\liminfT}{\xrightarrow[T \to \infty]{}}
\newcommand{\td}{\text{d}}
\newcommand{\softmax}{\text{softmax}}

\allowdisplaybreaks
\usepackage[small,compact]{titlesec}
\titlespacing{\section}{0pt}{1ex}{0.8ex}
\titlespacing{\subsection}{0pt}{0.5ex}{0ex}
\titlespacing{\subsubsection}{0pt}{0.2ex}{0ex}
\expandafter\def\expandafter\normalsize\expandafter{%
    \normalsize
    \setlength\abovedisplayskip{2pt}
    \setlength\belowdisplayskip{2pt}
    \setlength\abovedisplayshortskip{0pt}
    \setlength\belowdisplayshortskip{0pt}
}

\renewcommand{\baselinestretch}{0.99}

\usepackage{times}
\usepackage{graphicx} 

\usepackage{natbib}

\usepackage{algorithm}
\usepackage{algorithmic}

\usepackage{hyperref}

\usepackage[accepted]{icml2017}

\icmltitlerunning{Deep Bayesian Active Learning with Image Data}

\begin{document} 

\twocolumn[
\icmltitle{Deep Bayesian Active Learning with Image Data}




\begin{icmlauthorlist}
\icmlauthor{Yarin Gal}{cam,ati}
\icmlauthor{Riashat Islam}{cam}
\icmlauthor{Zoubin Ghahramani}{cam}
\end{icmlauthorlist}

\icmlaffiliation{cam}{University of Cambridge, UK}
\icmlaffiliation{ati}{The Alan Turing Institute, UK}

\icmlcorrespondingauthor{Yarin Gal}{yg279@cam.ac.uk}

\icmlkeywords{boring formatting information, machine learning, ICML}

\vskip 0.3in
]



\printAffiliationsAndNotice{}  

\begin{abstract} 
Even though active learning forms an important pillar of machine learning, deep learning tools are not prevalent within it. Deep learning poses several difficulties when used in an active learning setting. First, active learning (AL) methods generally rely on being able to learn and update models from small amounts of data. Recent advances in deep learning, on the other hand, are notorious for their dependence on large amounts of data. Second, many AL acquisition functions rely on model uncertainty, yet deep learning methods rarely represent such model uncertainty. 
In this paper we combine recent advances in Bayesian deep learning into the active learning framework in a practical way.
We develop an active learning framework for high dimensional data, a task which has been extremely challenging so far, with very sparse existing literature. 
Taking advantage of specialised models such as Bayesian convolutional neural networks, we demonstrate our active learning techniques with image data, obtaining a significant improvement on existing active learning approaches. We demonstrate this on both the MNIST dataset, as well as for skin cancer diagnosis from lesion images (ISIC2016 task). 
\end{abstract}

\section{Introduction}

A big challenge in many machine learning applications is obtaining labelled data. This can be a long, laborious, and costly process, often making the deployment of ML systems uneconomical. A framework where a system could learn from small amounts of data, and choose by itself what data it would like the user to label, would make machine learning much more widely applicable. Such frameworks for learning are referred to as \textit{active learning} \citep{cohn1996active} (also known as ``experiment design'' in the statistics literature), and have been used successfully in fields such as medical diagnosis, microbiology, and manufacturing \citep{Tong2001Active}. In active learning, a model is trained on a small amount of data (the initial training set), and an \textit{acquisition function} (often based on the model's \textit{uncertainty}) decides which data points to ask an external \textit{oracle} for a label. The acquisition function selects one or more points from a \textit{pool} of unlabelled data points, with the pool points lying outside of the training set. An oracle (often a human expert) labels the selected data points, these are added to the training set, and a new model is trained on the updated training set. This process is then repeated, with the training set increasing in size over time.
The advantage of such systems is that they often result in dramatic reductions in the amount of labelling required to train an ML system (and therefore cost and time).

Even though existing techniques for active learning have proven themselves useful in a variety of tasks, a major remaining challenge in active learning is its lack of scalability to high-dimensional data \citep{Tong2001Active}.
This data appears often in image form, with a physician classifying MRI scans to diagnose Alzheimer's for example \citep{marcus2010open}, or an expert clinician diagnosing skin cancer from dermoscopic lesion images.
To perform active learning, a model has to be able to learn from small amounts of data and represent its uncertainty over unseen data. This severely restricts the class of models that can be used within the active learning framework. As a result most approaches to active learning have focused on low dimensional problems \citep{Tong2001Active, hernandez2015probabilistic}, with only a handful of exceptions \citep{zhu2003combining, holub2008entropy, joshi2009multi} relying on kernel or graph-based approaches to handle high-dimensional data.

In recent years, with the increased availability of data in \textit{some} domains, attention within the machine learning community has shifted from small data problems to big data problems \citep{sundermeyer2012lstm, krizhevsky2012imagenet, kalchbrenner2013recurrent, sutskever2014sequence}. And with the increased interest in big data problems, new tools were developed and existing tools were refined for handling high dimensional data within such regimes. Deep learning, and convolutional neural networks (CNNs) \citep{rumelhart1985learning, lecun1989backpropagation} in particular, are an example of such tools. Originally developed in 1989 to parse handwritten zip codes, these tools have flourished and were adapted to a point where a CNN is able to beat a human on object recognition tasks (given enough training data) \citep{he2015delving}. New techniques such as dropout \citep{hinton2012improving, srivastava2014dropout} are used extensively to regularise these huge models, which often contain millions of parameters \citep{jozefowicz2016exploring}.
But even though active learning forms an important pillar of machine learning, deep learning tools are not prevalent within it. Deep learning poses several difficulties when used in an active learning setting. First, we have to be able to handle small amounts of data. Recent advances in deep learning, on the other hand, are notorious for their dependence on large amounts of data \cite{krizhevsky2012imagenet}. Second, many AL acquisition functions rely on model uncertainty. But in deep learning we rarely represent such model uncertainty. 

Relying on Bayesian approaches to deep learning, in this paper we combine recent advances in Bayesian deep learning into the active learning framework in a practical way.
We develop an active learning framework for high dimensional data, a task which has been extremely challenging so far with very sparse existing literature from the past 15 years \citep{zhu2003combining, li2013adaptive, holub2008entropy, joshi2009multi}. 
Taking advantage of specialised models such as Bayesian convolutional neural networks (BCNNs) \citep{Gal2016Bayesian, gal2016dropout}, we demonstrate our active learning techniques with image data. Using a small model, our system is able to achieve 5\% test error on MNIST with only 295 labelled images without relying on unlabelled data (in comparison, 835 labelled images are needed to achieve 5\% test error using random sampling -- 
requiring an expert to label more than twice as many images to achieve the same accuracy), and achieves 1.64\% test error with 1000 labelled images. This is in comparison to 2.40\% test error of DGN \citep{kingma2014semi} or 1.53\% test error of the Ladder Network $\Gamma$-model \citep{rasmus2015semi}, both semi-supervised learning techniques which additionally use the entire unlabelled training set.
Finally, we study a real-world application by diagnosing melanoma (skin cancer) from a small number of lesion images by fine-tuning the VGG16 convolutional neural network \citep{Simonyan15} on the ISIC 2016 dataset \citep{gutman2016skin}. 

\section{Related Research}

Past attempts at active learning of image data have concentrated on kernel based methods.
Using ideas from previous research in active learning of low dimensional data \citep{Tong2001Active}, \citet{joshi2009multi} used ``margin-based uncertainty'' and extracted probabilistic outputs from support vector machines (SVM) \citep{cortes1995support}.
They used linear, polynomial, and Radial Basis Function (RBF) kernels on the raw images, picking the kernel that gave best classification accuracy.
Analogously to SVM approaches, \citet{li2013adaptive} used Gaussian processes (GPs) with RBF kernels to get model uncertainty. However \citet{li2013adaptive} fed low dimensional features (such as SIFT features) to their RBF kernel.
Lastly, making use of unlabelled data as well, \citet{zhu2003combining} acquire points using a Gaussian random field model, evaluating an RBF kernel over raw images. We compare to this last technique and explain it in more detail below. 

Other related work includes semi-supervised learning of image data \citep{weston2012deep, kingma2014semi, rasmus2015semi}. 
In semi-supervised learning a model is given a fixed set of labelled data, and a fixed set of unlabelled data. The model can use the unlabelled data to learn about the distribution of the inputs, in the hopes that this information will aid in learning from the small labelled set as well. Although the learning paradigm is fairly different from active learning, this research forms the closest modern literature to active learning of image data. We will compare to these techniques below as well, in section \ref{sect:SS_exp}.


\section{Bayesian Convolutional Neural Networks}

In this paper we concentrate on high dimensional \textit{image} data, and need a model able to represent prediction uncertainty on such data. Existing approaches such as \citep{zhu2003combining, li2013adaptive, joshi2009multi} rely on kernel methods, and feed image pairs through linear, polynomial, and RBF kernels to capture image similarity as an input to an SVM for example.
In contrast, we rely on specialised models for image data, and in particular on convolutional neural networks (CNNs) \citep{rumelhart1985learning, lecun1989backpropagation}. 
Unlike the kernels above, which cannot capture spatial information in the input image, CNNs are designed to use this spatial information, and have been used successfully to achieve state-of-the-art results \citep{krizhevsky2012imagenet}.
To perform active learning with image data we make use of the Bayesian equivalent of CNNs, proposed in \citep{Gal2016Bayesian}\footnote{As far as we are aware, there are no other tools in current literature that offer model uncertainty in specialised models for image data, which perform as well as CNNs.}.
These Bayesian CNNs are CNNs with prior probability distributions placed over a set of 
model parameters $\bo = \{W_1, ..., W_L\}$:
$$
\bo \sim p(\bo),
$$
with for example a standard Gaussian prior $p(\bo)$.
We further define a likelihood model
$$
p(y=c | \x, \bo) = \softmax(\f^\bo(\x))
$$
for the case of classification, or a Gaussian likelihood for the case of regression, with $\f^\bo(\x)$ model output (with parameters $\bo$).

To perform approximate inference in the Bayesian CNN model we make use of stochastic regularisation techniques such as dropout \citep{hinton2012improving, srivastava2014dropout}, originally used to regularise these models. As shown in \citep{gal2016dropout, Gal2016Uncertainty} dropout and various other stochastic regularisation techniques can be used to perform practical approximate inference in complex deep models. Inference is done by training a model with dropout before every weight layer, and by performing dropout at test time as well to sample from the approximate posterior (stochastic forward passes, referred to as \textit{MC dropout}).

More formally, this approach is equivalent to performing approximate variational inference where we find a distribution $q_\theta^*(\bo)$ in a tractable family which minimises the Kullback-Leibler (KL) divergence to the true model posterior $p(\bo | \D_\train)$ given a training set $\D_\train$. Dropout can be interpreted as a variational Bayesian approximation, where the approximating distribution is a mixture of two Gaussians with small variances and the mean of one of the Gaussians is fixed at zero. The uncertainty in the weights induces prediction uncertainty by marginalising over the approximate posterior using Monte Carlo integration:
\begin{align*}
p(y=c | \x, \D_\train) &= \int p(y=c | \x, \bo) p(\bo | \D_\train) \td \bo \\
&\approx 
\int p(y=c | \x, \bo) q_\theta^*(\bo) \td \bo \\
&\approx 
\frac{1}{T} \sum_{t=1}^T p(y=c | \x, \boh_t)
\end{align*}
with $\boh_t \sim q_\theta^*(\bo)$, where $q_\theta(\bo)$ is the Dropout distribution \citep{Gal2016Uncertainty}.

Bayesian CNNs work well with small amounts of data \citep{Gal2016Bayesian}, and possess uncertainty information that can be used with existing acquisition functions \citep{Gal2016Uncertainty}. Such acquisition functions for the case of classification are discussed next.


\section{Acquisition Functions and their Approximations}
\label{sec:acq}

Given a model $\mathcal{M}$, pool data $\mathcal{D}_\text{pool}$, and inputs $x \in \mathcal{D}_\text{pool}$, an acquisition function $a(x, \mathcal{M})$ is a function of $x$ that the AL system uses to decide where to query next:
\begin{align*}
x^* = \text{argmax}_{x \in \D_\text{pool}} a(x, \mathcal{M}).
\end{align*}
We next explore various acquisition functions appropriate for our image data setting, and develop tractable approximations for us to use with our Bayesian CNNs.
In tasks involving regression we often use the predictive variance or a quantity derived from this for our acquisition function (although we still need to be careful to query from informative areas rather than querying noise). For example, we might look for images with high predictive variance and choose those to ask an expert to label -- in the hope that these will decrease model uncertainty. 
However, many tasks involving image data are often phrased as classification problems. 
For classification, several acquisition functions are available:
\begin{enumerate}
\item
Choose pool points that maximise the predictive entropy (\textit{Max Entropy}, \citep{shannon1948mathematical})
\begin{align*}
&\bbH [y | \x, \D_\train] := \\
&\quad - \sum_c p(y=c | \x, \D_\train) \log p(y=c | \x, \D_\train).
\end{align*}
\item
Choose pool points that are expected to maximise the information gained about the model parameters, i.e.\
maximise the mutual information between predictions and model posterior 
(\textit{BALD}, \citep{houlsby2011bayesian})
$$
\bbI[y, \bo | \x, \D_\train] =
\bbH [y | \x, \D_\train] - \E_{p(\bo | \D_\train)} \big[ \bbH[y | \x, \bo] \big] 
$$
with $\bo$ the model parameters (here $\bbH[y | \x, \bo] $ is the entropy of $y$ given model weights $\bo$).
Points that maximise this acquisition function are points on which the model is uncertain on average, but there exist model parameters that produce disagreeing predictions with high certainty. This is equivalent to points with high variance in the input to the softmax layer (the logits) -- thus each stochastic forward pass through the model would have the highest probability assigned to a different class. 
\item
Maximise the \textit{Variation Ratios} \citep{freeman1965elementary}
$$
\varration[\x] := 1 - \max_y p(y | x, \D_\train)
$$
Like \textit{Max Entropy}, \textit{Variation Ratios} measures lack of confidence.
\item
Maximise mean standard deviation (\textit{Mean STD}) \citep{Kampffmeyer2016Semantic, kendall2015bayesian}
\begin{gather*}
\sigma_c = \sqrt{\E_{q(\bo)}[p(y=c | \x, \bo)^2] - \E_{q(\bo)}[p(y=c | \x, \bo)]^2} \\
\sigma(\x) = \frac{1}{C} \sum_c \sigma_c
\end{gather*}
averaged over all $c$ classes $\x$ can take. Compared to the above acquisition functions, this is more of an ad-hoc technique used in recent literature.
\item
\textit{Random} acquisition (baseline): 
$
a(\x) = \text{unif}()
$
with $\text{unif}()$ a function returning a draw from a uniform distribution over the interval $[0, 1]$. Using this acquisition function is equivalent to choosing a point uniformly at random from the pool.
\end{enumerate}

These acquisition functions and their properties are discussed in more detail in \citep[pp.\ 48--52]{Gal2016Uncertainty}.

We can approximate each of these acquisition functions using our approximate distribution $q_\theta^*(\bo)$. For BALD, for example, we can write the acquisition function as follows:
\begin{align*}
&\bbI[y, \bo | \x, \D_\train] 
:=
\bbH [y | \x, \D_\train] - \E_{p(\bo | \D_\train)} \big[ \bbH[y | \x, \bo] \big] 
\\
&\quad =
- \sum_c p(y=c | \x, \D_\train) \log p(y=c | \x, \D_\train) 
\\
&\quad \qquad + \E_{p(\bo | \D_\train)} \bigg[ 
\sum_c p(y=c | \x, \bo) \log p(y=c | \x, \bo) 
\bigg],
\end{align*}
with $c$ the possible classes $y$ can take.
$\bbI[y, \bo | \x, \D_\train]$ can be approximated in our setting using the identity $p(y=c | \x, \D_\train) = \int p(y=c | \x, \bo) p(\bo | \D_\train) \td \bo$:
\begin{align*}
&\bbI[y, \bo | \x, \D_\train] =\\
&\quad - \sum_c \int p(y=c | \x, \bo) p(\bo | \D_\train) \td \bo \\
&\quad \qquad \qquad \cdot \log \int p(y=c | \x, \bo) p(\bo | \D_\train) \td \bo 
\\
&\quad + \E_{p(\bo | \D_\train)} \bigg[ 
\sum_c p(y=c | \x, \bo) \log p(y=c | \x, \bo) 
\bigg].
\end{align*}
Swapping the posterior $p(\bo | \D_\train)$ with our approximate posterior $q_\theta^*(\bo)$, and through MC sampling, we then have:
\begin{align*}
&\approx
- \sum_c \int p(y=c | \x, \bo) q_\theta^*(\bo) \td \bo \\
& \qquad \qquad \qquad \cdot \log \int p(y=c | \x, \bo) q_\theta^*(\bo) \td \bo 
\\
&\quad + \E_{q_\theta^*(\bo)} \bigg[ 
\sum_c p(y=c | \x, \bo) \log p(y=c | \x, \bo) 
\bigg] \\
&\approx - \sum_c \bigg(\frac{1}{T} \sum_t \ph_c^t \bigg) \log \bigg(\frac{1}{T} \sum_t \ph_c^t \bigg)
\\
&\qquad + \frac{1}{T} \sum_{c,t} \ph_c^t \log \ph_c^t =: \widehat{\bbI}[y, \bo | \x, \D_\train]
\end{align*}
defining our approximation, with $\ph_c^t$ the probability of input $\x$ with model parameters $\boh_t \sim q_\theta^*(\bo)$ to take class $c$:
\begin{align*}
\bph^t = [\ph_1^t, ..., \ph_C^t] = \softmax(\f^{\boh_t}(\x)).
\end{align*}
We then have
\begin{align*}
\widehat{\bbI}[y, \bo | \x, \D_\train]
&\liminfT
\bbH [y | \x, q_{\theta}^*] - \E_{q_\theta^*(\bo)} \big[ \bbH[y | \x, \bo] \big] 
\\
&\approx 
\bbI[y, \bo | \x, \D_\train],
\end{align*}
resulting in a computationally tractable estimator approximating the BALD acquisition function.
The other acquisition functions can be approximated similarly.

In the next section we will experiment with these acquisition functions and assess them empirically. 
These will be compared to the baseline acquisition function which uniformly acquires new data points from the pool set at random, and to various other techniques for active learning of image data and semi-supervised learning. This is followed by a real-world case study using cancer diagnosis.

\section{Active Learning with Bayesian Convolutional Neural Networks}

We study the proposed technique for active learning of image data. We compare the various acquisition functions relying on Bayesian CNN uncertainty with a simple image classification benchmark. We then study the importance of model uncertainty by evaluating the same acquisition functions with a deterministic CNN. This is followed by a comparison to a current technique for active learning with image data, which relies on SVMs.
We follow with a comparison to the closest modern models to our active learning with image data -- semi-supervised techniques with image data. These semi-supervised techniques have access to much more data (the unlabelled data) than our active learning models, yet we still perform in comparable terms to them. 
Finally, we demonstrate the proposed methodology with a real world application of skin cancer diagnosis from a small number of lesion images, relying on fine-tuning of a large CNN model.

\subsection{Comparison of various acquisition functions}
We next study all acquisition functions above with our Bayesian CNN trained on the MNIST dataset \citep{lecun1998mnist}.
All acquisition functions are assessed with the same model structure: convolution-relu-convolution-relu-max pooling-dropout-dense-relu-dropout-dense-softmax, with 32 convolution kernels, 4x4 kernel size, 2x2 pooling, dense layer with 128 units, and dropout probabilities 0.25 and 0.5 (following the example Keras MNIST CNN implementation \citep{keras2015}).

\begin{figure}[b!]
\center
\includegraphics[width=\linewidth]{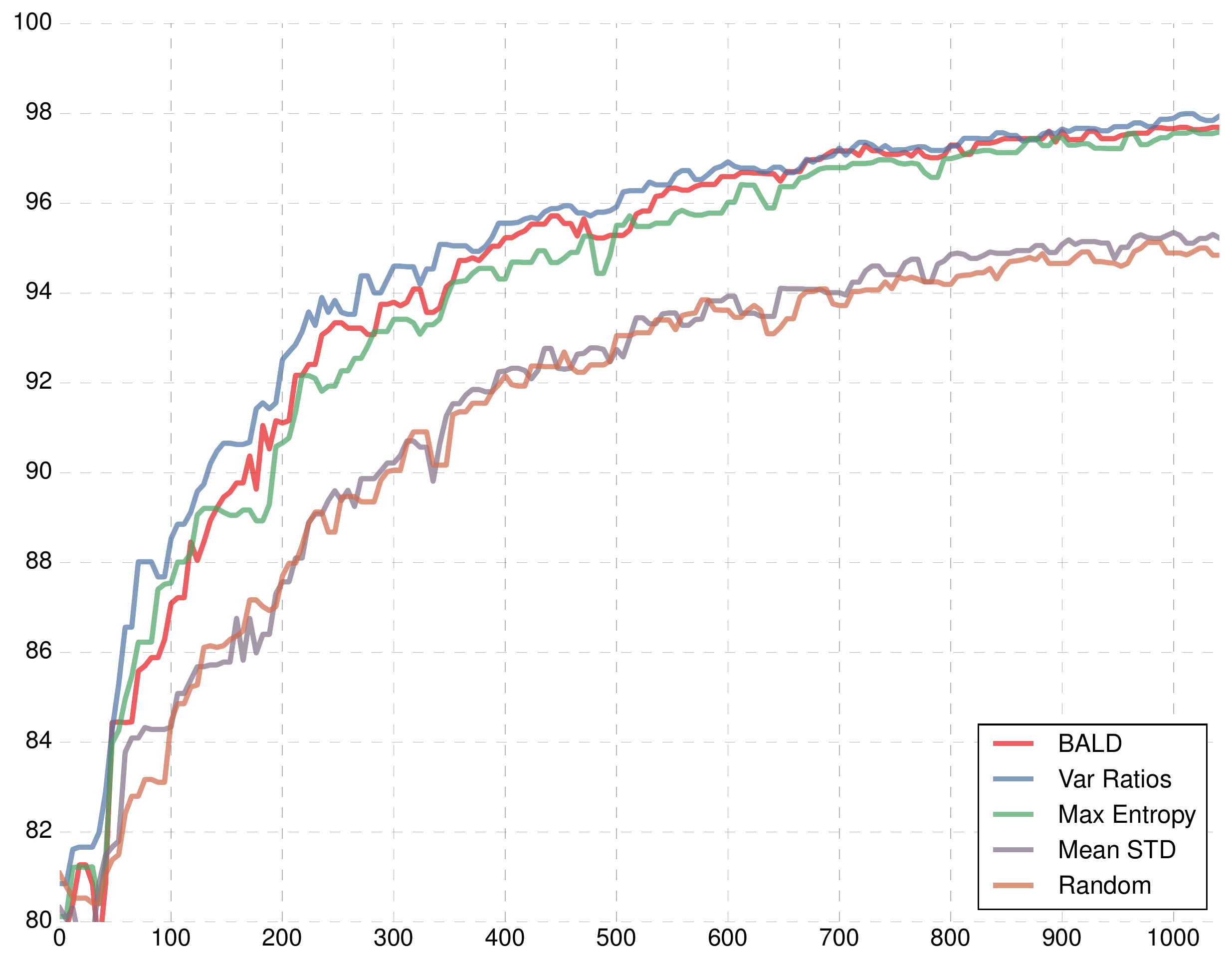}
\caption{
\textbf{MNIST test accuracy as a function of number of acquired images from the pool set} (up to 1000 images, using validation set size 100, and averaged over 3 repetitions).
Four acquisition functions (\textit{BALD}, \textit{Variation Ratios}, \textit{Max Entropy}, and \textit{Mean STD}) are evaluated and compared to a \textit{Random} acquisition function.}
\label{fig:active}
\end{figure}

\begin{figure*}[t!]
\center
\begin{subfigure}[t]{0.3\linewidth}
\includegraphics[width=\linewidth]{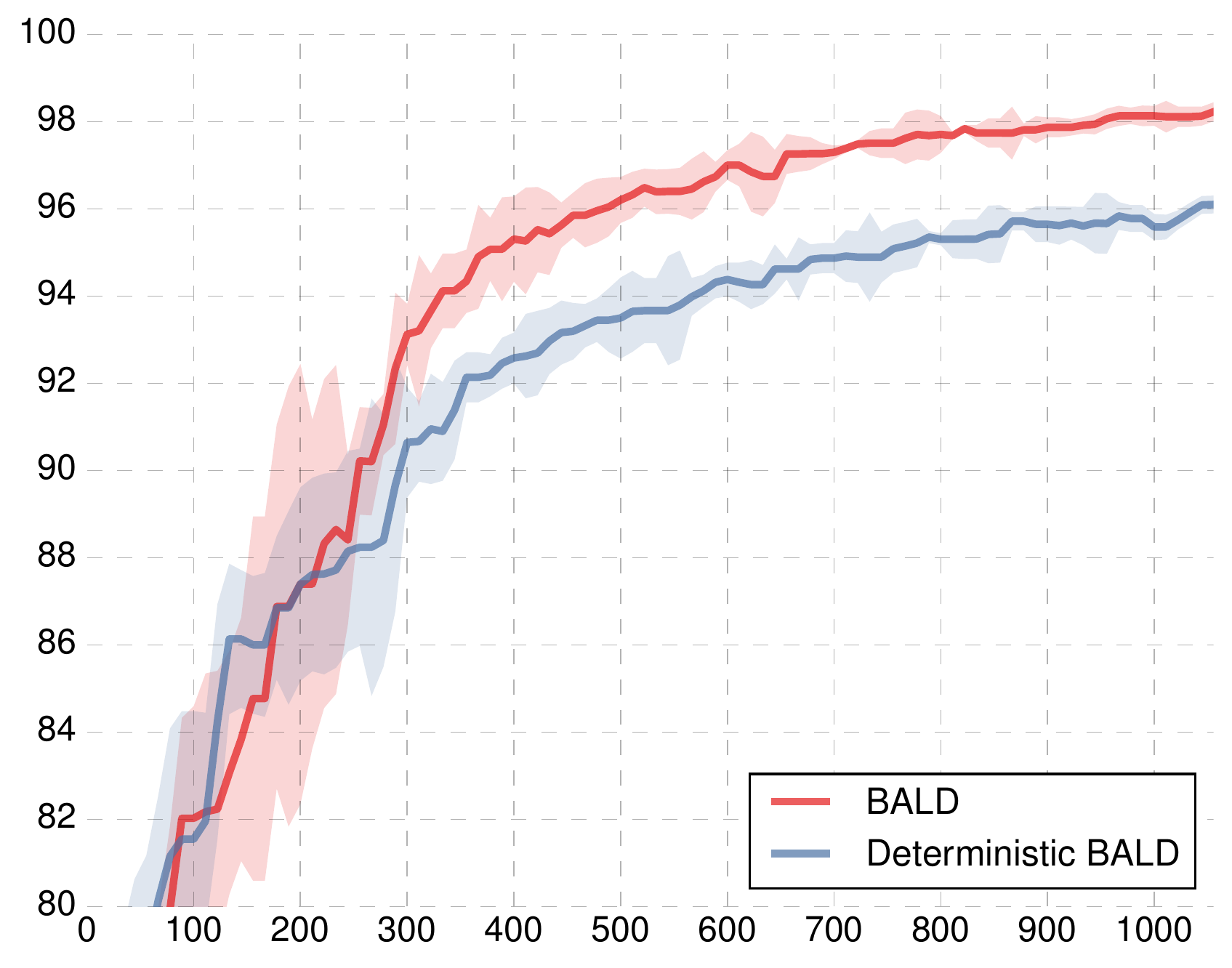}
\caption{BALD}
\end{subfigure}
\hspace{5mm}
\begin{subfigure}[t]{0.3\linewidth}
\includegraphics[width=\linewidth]{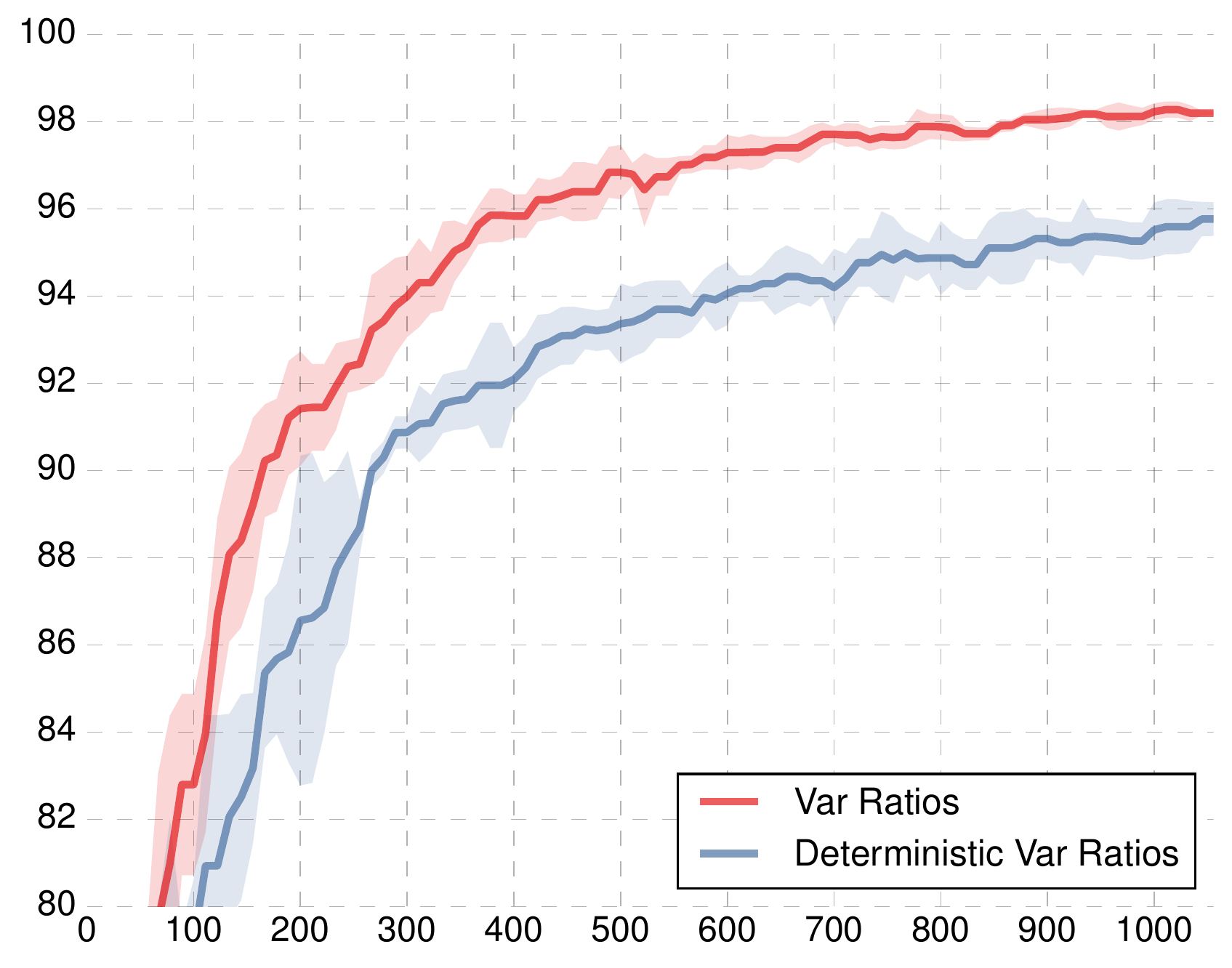}
\caption{Var Ratios}
\end{subfigure}
\hspace{5mm}
\begin{subfigure}[t]{0.3\linewidth}
\includegraphics[width=\linewidth]{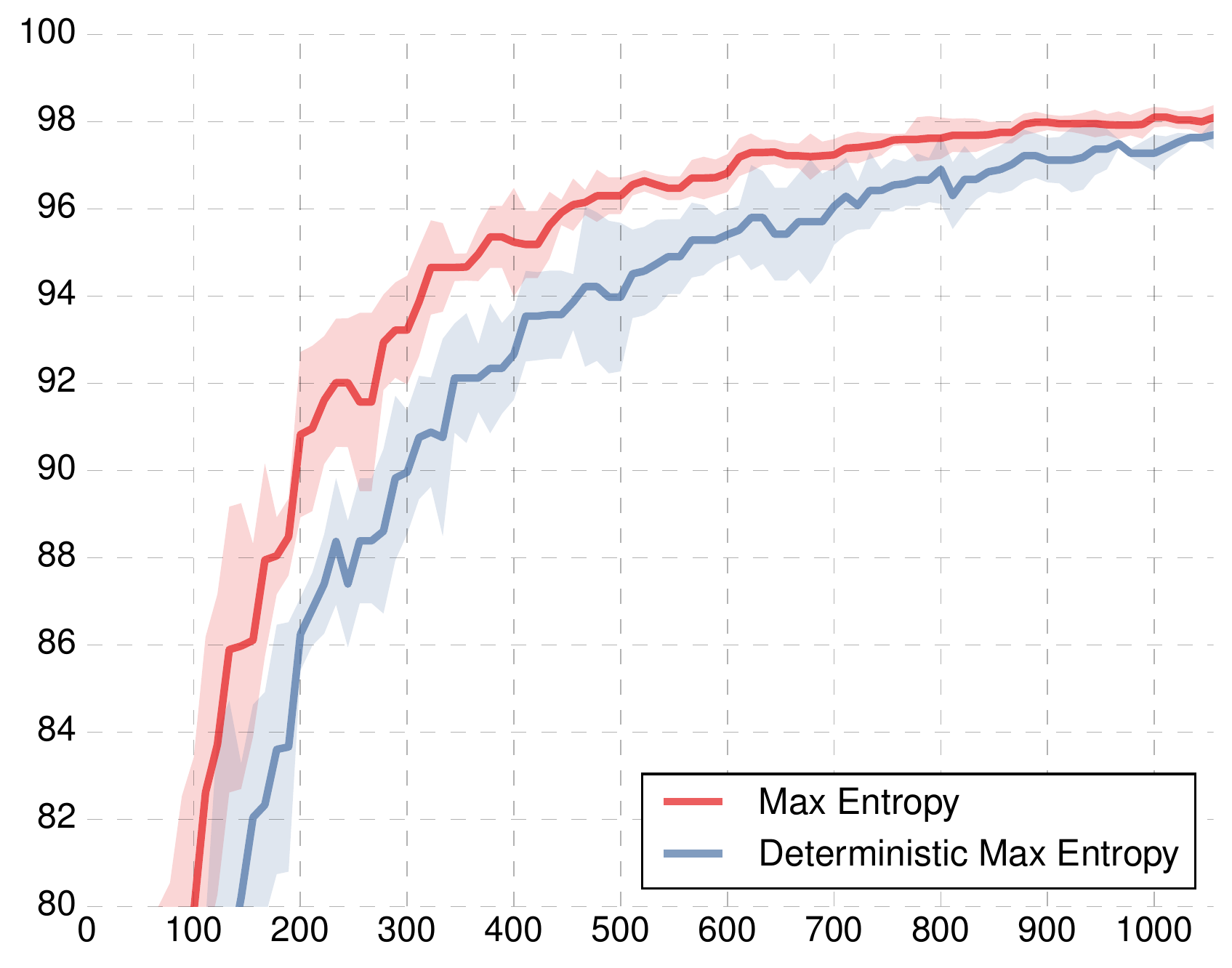}
\caption{Max Entropy}
\end{subfigure}
\caption{Test accuracy as a function of number of acquired images for various acquisition functions, using both a \textbf{Bayesian CNN (red) and a deterministic CNN (blue)}.}
\label{fig:deterministic}
\end{figure*}

All models are trained on the MNIST dataset with a (random but balanced) initial training set of 20 data points, and a validation set of 100 points on which we optimise the weight decay (this is a realistic validation set size, in comparison to the standard validation set size of 5K used in similar applications such as semi-supervised learning on MNIST). We further use the standard test set of 10K points, and the rest of the points are used as a pool set. 
The test error of each model and each acquisition function was assessed after each acquisition, using the dropout approximation at test time. To decide what data points to acquire though we used MC dropout following the derivations above.
We repeated the acquisition process 100 times, each time acquiring the 10 points that maximised the acquisition function over the pool set. Each experiment was repeated three times and the results averaged (the standard deviation for the three repetitions is shown below)
\footnote{The code for these experiments is available at \texttt{\url{http://mlg.eng.cam.ac.uk/yarin/publications.html\#Gal2016Active}}.}.

We compared the acquisition functions BALD, Variation Ratios, Max Entropy, Mean STD, and the baseline Random. We found Random and Mean STD to under-perform compared to BALD, Variation Ratios, and Max Entropy (figure \ref{fig:active}). 
The Variation Ratios acquisition function seems to obtain slightly better accuracy faster than BALD and Max Entropy. 
It is interesting that Mean STD seems to perform similarly to Random -- which samples points at random from the pool set.

Lastly, in table \ref{table:model-error-num-acqs} we give the number of acquisition steps needed to get to test errors of 5\% and 10\%. As can be seen, BALD, Variation Ratios, and Max Entropy attain a small test error with much fewer acquisitions than Mean STD and Random. This table demonstrates the importance of data efficiency -- an expert using the Variation Ratios model for example would have to label less than half the number of images she would have had to label had she acquired new images at random.

\begin{table}[h]
    \def\arraystretch{1.25}
    \setlength{\tabcolsep}{2pt}
\center
\begin{tabular}{c|ccccc}
\% error & BALD & Var Ratios & Max Ent & Mean STD & Random \\ 
\midrule
10\% & 145 & 120 & 165 & 230 & 255 \\ 
5\% & 335 & 295 & 355 & 695 & 835 \\ 
\bottomrule
\end{tabular} 

\vspace{2mm}
\caption{
Number of acquired images to get to model error of \% on MNIST.
}
\label{table:model-error-num-acqs}
\end{table}

\subsection{Importance of model uncertainty}

We assess the importance of model uncertainty in our Bayesian CNN by evaluating three of the acquisition functions (BALD, Variation Ratios, and Max Entropy) with a deterministic CNN. Much like the Bayesian CNN, the deterministic CNN produces a probability vector which can be used with the acquisition functions of \S\ref{sec:acq} (formally, by setting $q_\theta^*(\bo) = \delta(\bo - \theta)$ to be a point mass at the location of the model parameters $\theta$).
Such deterministic models can capture \textit{aleatoric uncertainty} -- the noise in the data -- but cannot capture \textit{epistemic uncertainty} -- the uncertainty over the parameters of the CNN, which we try to minimise during active learning.
The models in this experiment still use dropout, but for regularisation only (i.e.\ we do not perform MC dropout at test time).

A comparison of the Bayesian models to the deterministic models for the BALD, Variation Ratios, and Max Entropy acquisition functions is given in fig.\ \ref{fig:deterministic}. 
The Bayesian models, propagating uncertainty throughout the model, attain higher accuracy early on, and converge to a higher accuracy overall. 
This demonstrates that the uncertainty propagated throughout the Bayesian models has a significant effect on the models' measure of their confidence.

\subsection{Comparison to current active learning techniques with image data}

We next compare to a method in the sparse existing literature of active learning with image data, concentrating on \citep{zhu2003combining} which relies on a kernel method and further leverages the unlabelled images (which will be discussed in more detail in the next section). 
\citet{zhu2003combining} evaluate an RBF kernel over the raw images to get a similarity graph which can be used to share information about the unlabelled data. Active learning is then performed by greedily selecting unlabelled images to be labelled, such that an estimate to the expected classification error is minimised. This will be referred to as \textit{MBR}.

\begin{figure*}[t!]
\center
\begin{minipage}[t]{0.9\linewidth}
\includegraphics[width=\linewidth]{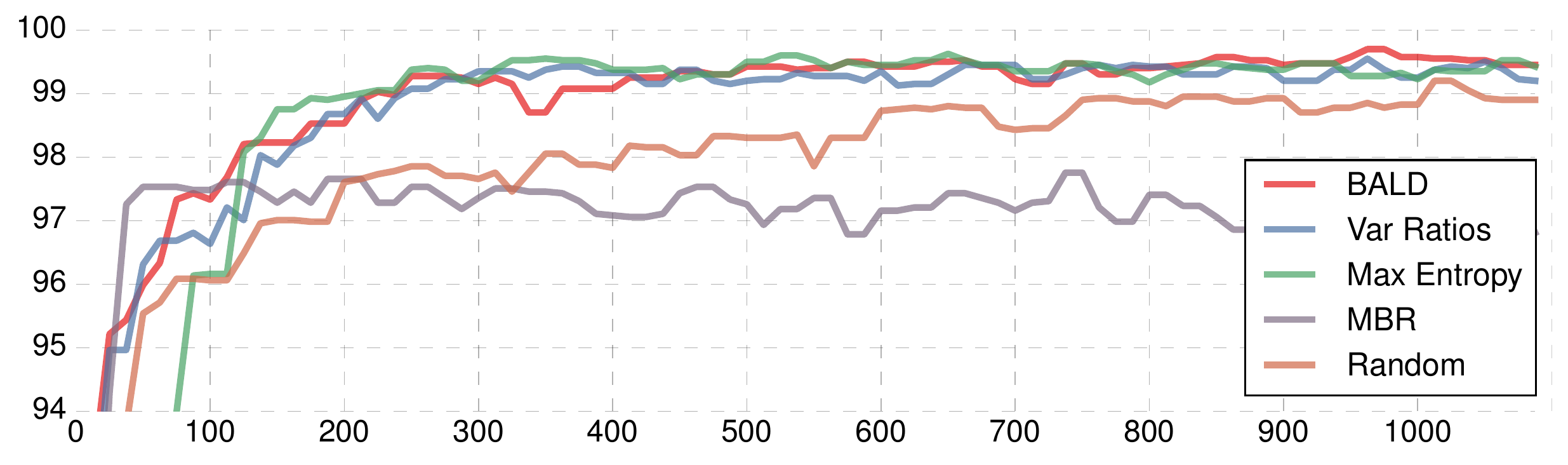}
\caption{
MNIST test accuracy (two digit classification) as a function of number acquired images, compared to a current technique for active learning of image data:
MBR \citep{zhu2003combining}.
}
\label{fig:mbr}
\end{minipage}
\end{figure*}

MBR was formulated for the binary classification case, hence we compared MBR to the acquisition functions BALD, Variation Ratios, Max Entropy, and Random on a binary classification task (two digits from the MNIST dataset). 
Classification accuracy is shown in fig.\ \ref{fig:mbr}. Note that even a random acquisition function, when coupled with a CNN (a specialised model for image data) outperforms MBR which relies on an RBF kernel. We further experimented with a CNN version for MBR where we replaced the RBF kernel with a CNN. It is interesting to note that this did not give improved results.

\subsection{Comparison to semi-supervised learning}
\label{sect:SS_exp}

We continue with a comparison to the closest models (in modern literature) to our active learning with image data: semi-supervised learning with image data.
In \textit{semi-supervised learning} a model is given a fixed set of labelled data, and a fixed set of unlabelled data. The model can use the unlabelled dataset to learn about the distribution of the inputs, in the hopes that this information will aid in learning the mapping to the outputs as well. 
Several semi-supervised models for image data have been suggested in recent years \citep{weston2012deep, kingma2014semi, rasmus2015semi}, models which have set benchmarks on MNIST given a small number of \textit{labelled} images (1000 random images). These models make further use of a (very) large unlabelled set of 49K images, and a large validation set of 5K-10K \textit{labelled images} to tune model hyper-parameters and model structure \citep{rasmus2015semi}.
These models have access to much more data than our active learning models, but we still compare to them as they are the most relevant models in the field given the constraint of small amounts of \textit{labelled} data.

Test error for our active learning models with various acquisition functions (after the acquisition of 1000 training points), as well as the semi-supervised models, is given in table \ref{table:model-error}. In this experiment, to be comparable to the other techniques, we use a validation set of 5K points. Our model attains similar performance to that of the semi-supervised models (although note that we use a fairly small model compared to \citep{rasmus2015semi} for example). \citet{rasmus2015semi}'s ladder network (full) attains error 0.84\% with 1000 labelled images and 59,000 unlabelled images. However, \citep{rasmus2015semi}'s $\Gamma$-model architecture is more directly comparable to ours. The $\Gamma$-model attains 1.53\% error, compared to 1.64\% error of our Var Ratio acquisition function which relies on no additional unlabelled data.

\begin{center}
\begin{table}[b!]
\begin{tabular}{l c}
\toprule
\textbf{Technique} & Test error \\
\midrule
\textbf{Semi-supervised:} \\
\midrule
Semi-sup. Embedding 
\citep{weston2012deep} & 5.73\% \\
Transductive SVM 
\citep{weston2012deep} & 5.38\% \\
MTC \citep{rifai2011manifold} & 3.64\% \\
Pseudo-label \citep{lee2013pseudo} & 3.46\% \\
AtlasRBF \citep{pitelis2014semi} & 3.68\% \\
DGN \citep{kingma2014semi} & 2.40\% \\
Virtual Adversarial 
\citep{miyato2015distributional} & 1.32\% \\
Ladder Network ($\Gamma$-model)
\citep{rasmus2015semi} \hspace{-6mm} & 1.53\% \\
Ladder Network (full)
\citep{rasmus2015semi} & 0.84\% \\
\midrule
\textbf{Active learning with} \\
\textbf{various  acquisitions}: \\
\midrule
Random & 4.66\% \\
BALD & 1.80\% \\
Max Entropy & 1.74\% \\
Var Ratios & 1.64\% \\
\bottomrule
\end{tabular}
\caption{
\textbf{Test error on MNIST with 1000 labelled training samples, compared to semi-supervised techniques.}
Active learning has access to only the 1000 acquired images. Semi-supervised further has access to the remaining images with no labels. Following existing research we use a large validation set of size 5000.
}
\label{table:model-error}
\end{table}
\end{center}

\subsection{Cancer diagnosis from lesion image data}

\begin{figure*}[t!]
\center
\begin{subfigure}[t]{0.2\linewidth}
\includegraphics[width=\linewidth]{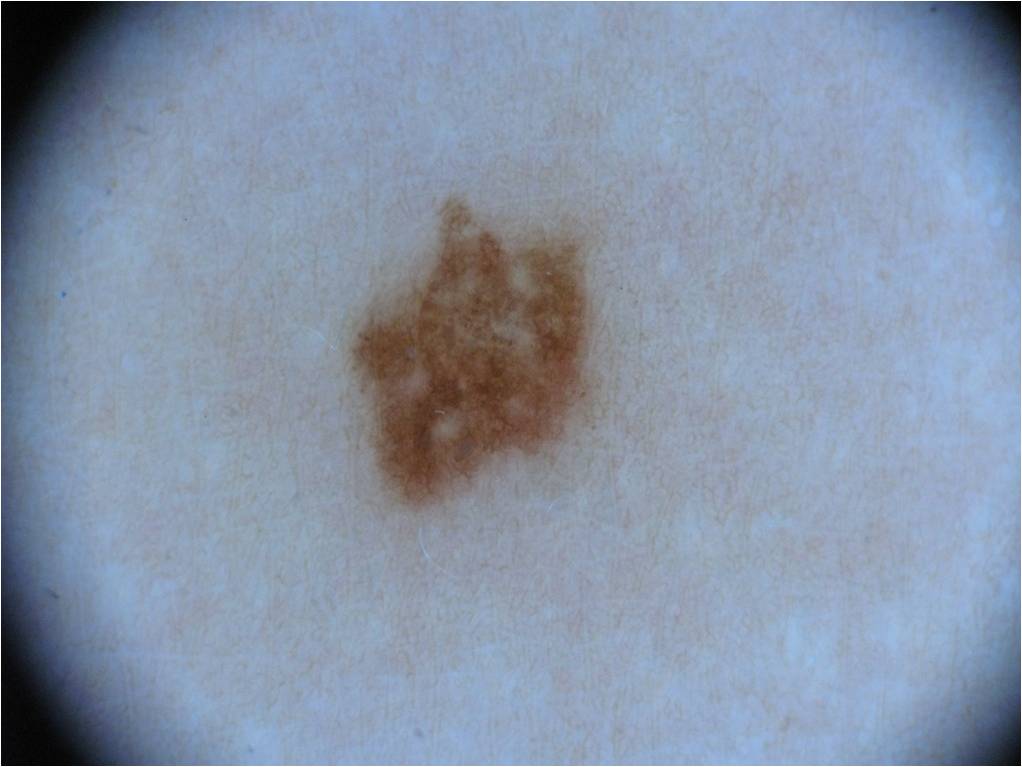}
\end{subfigure}
\hspace{5mm}
\begin{subfigure}[t]{0.2\linewidth}
\includegraphics[width=\linewidth]{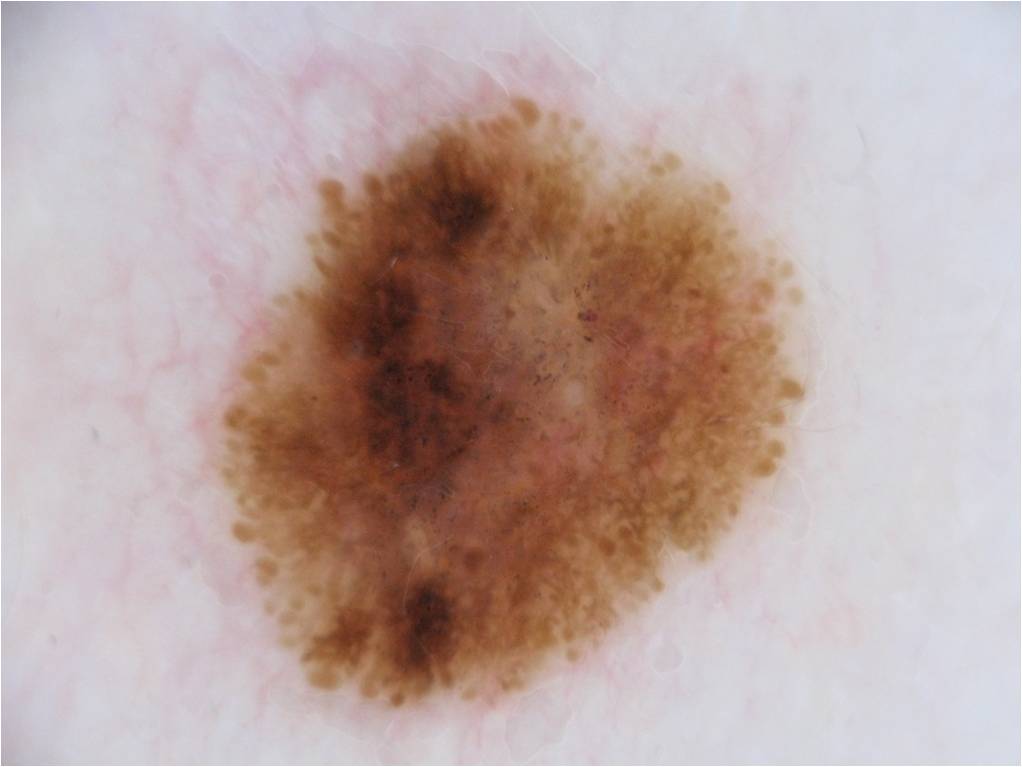}
\end{subfigure}
\hspace{5mm}
\begin{subfigure}[t]{0.2\linewidth}
\includegraphics[width=\linewidth]{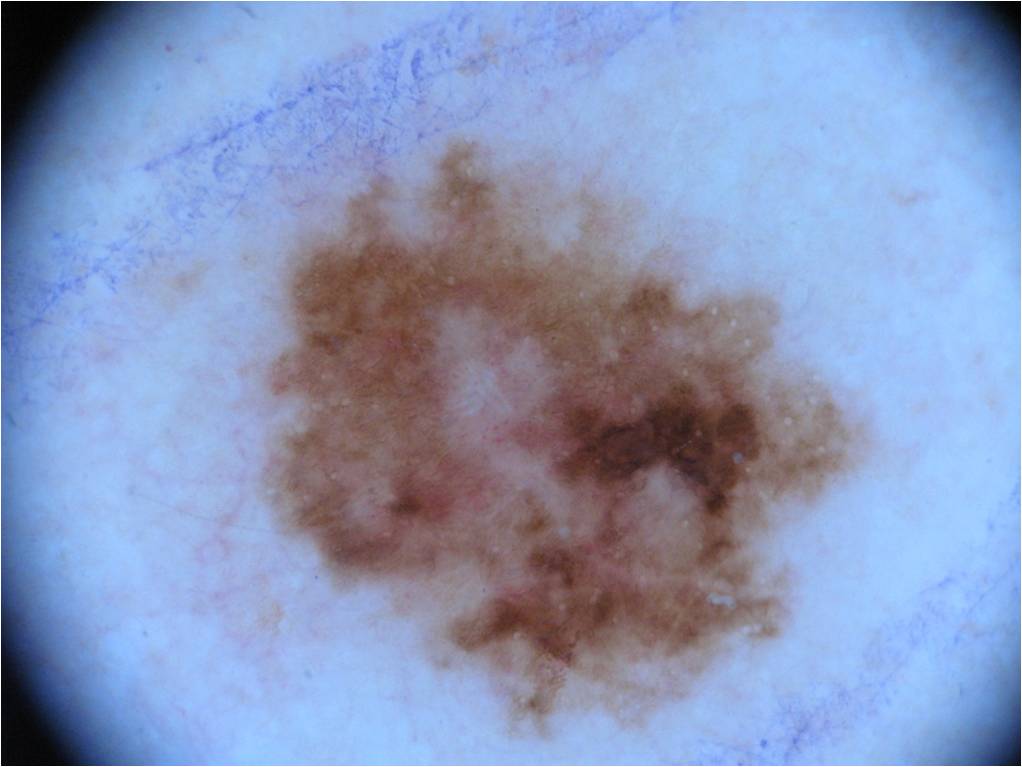}
\end{subfigure}
\hspace{5mm}
\begin{subfigure}[t]{0.2\linewidth}
\includegraphics[width=\linewidth]{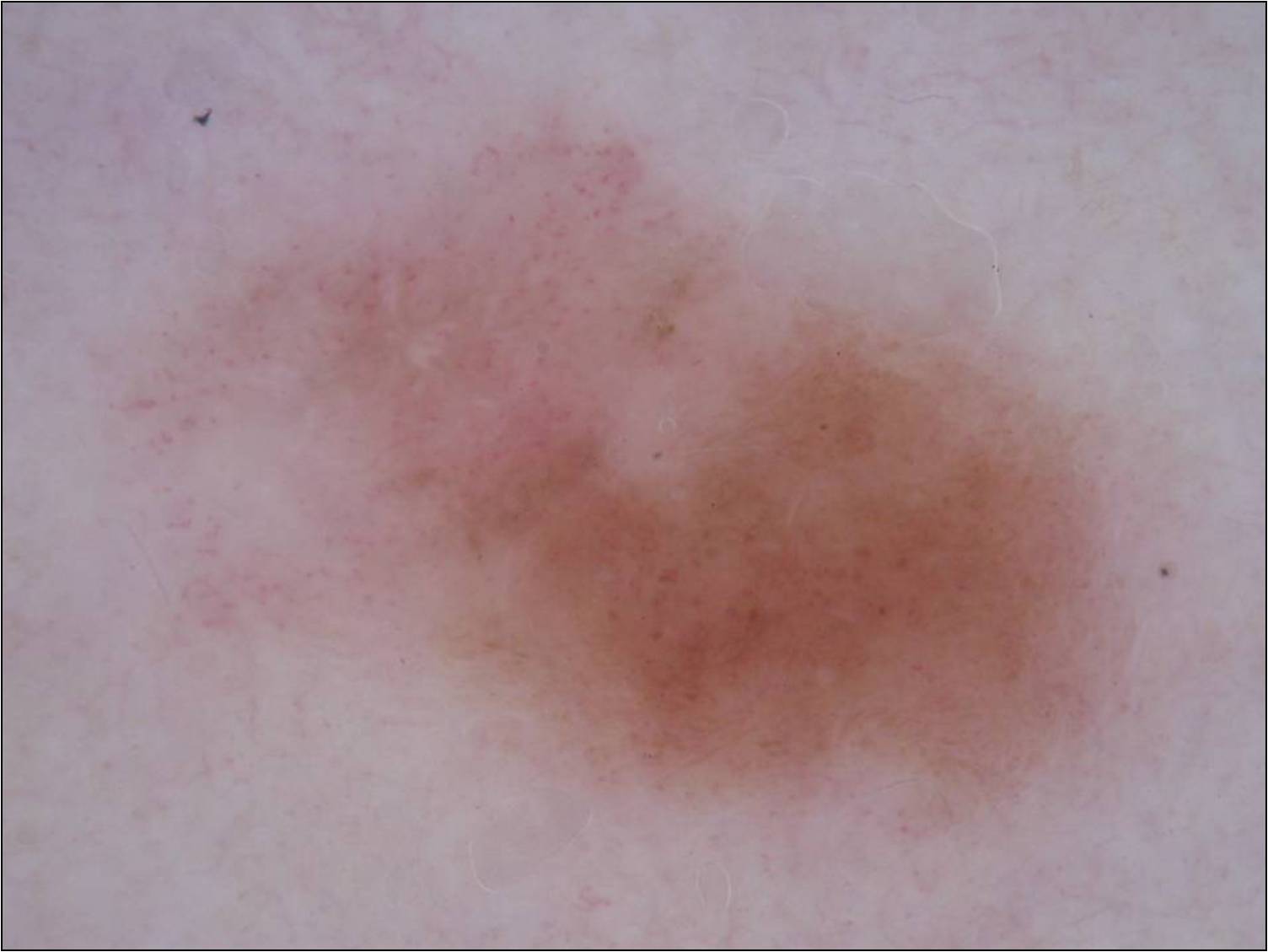}
\end{subfigure}

\vspace{-2mm}
\caption{Skin cancer (melanoma) example lesions from the ISIC 2016 melanoma diagnosis dataset. The two lesions on the left are benign (non-cancerous), while the two lesions on the right are malignant (cancerous).}
\label{fig:melanoma_photo}
\end{figure*}

We finish by assessing the proposed technique with a real world test case. We experiment with melanoma (skin cancer) diagnosis from dermoscopic lesion images. In this task we are given image data of skin segments, of both malignant (cancerous) as well as benign (non-cancerous) lesions. Our task is to classify the images as malignant or benign (an example is shown in fig.\ \ref{fig:melanoma_photo}). The data used is the ISIC Archive \citep{gutman2016skin}. This dataset was collected in order to provide a ``large public repository of expertly annotated high quality skin images'' to provide clinical support in the identification of skin cancer, and to develop algorithms for skin cancer diagnosis.
Specifically, we use the training data of the ``ISBI 2016: Skin Lesion Analysis Towards Melanoma Detection -- Part 3B: Segmented Lesion Classification'' task.
The data contains 900 dermoscopic lesion images in JPEG format with EXIF tags removed. Malignancy diagnosis for these lesions was obtained from expert consensus and pathology report information. The data contains lesion segmentation as well, which we did not use.

For our model we replicate the model of \citep{Agarwal2016}. This model achieved second place in the ``Part 3B: Segmented Lesion Classification'' task, with its code open-sourced. The model relies on data augmentation of the positive examples (flipping the lesions vertically and horizontally), and fine-tunes the VGG16 CNN model \citep{Simonyan15} (i.e.\ optimises a pre-trained model with a small learning rate). The VGG16 model was pre-trained on ImageNet \citep{deng2009imagenet}. The top layer of the model (1000 logits) was removed and replaced with a 2 dimensional output (for our classification task of malignant/benign). Preceding the last layer are two fully connected layers of size 4096, each one followed by a dropout layer with dropout probability 0.5. This architecture seems to provide good uncertainty estimates as observed before \citep{kendall2015bayesian,Gal2016Bayesian}. 

The data is unbalanced, containing 727 negative (benign) examples, and 173 positive (malignant) examples (20\% positive examples).
Since the data is so small, to assess model performance reliably we have to take a large balanced test set. We randomly partition the data, and set aside 100 negative and 100 positive examples. All our experiments are performed on two different random splits -- since even a test set size of 200 gives very different accuracy with different random splits. Note that \textit{on each such random split} we repeat our experiments three times and average the results with respect to the fixed test set.

\begin{figure*}[t!]
\center
\begin{subfigure}[t]{0.4\linewidth}
\includegraphics[width=\linewidth]{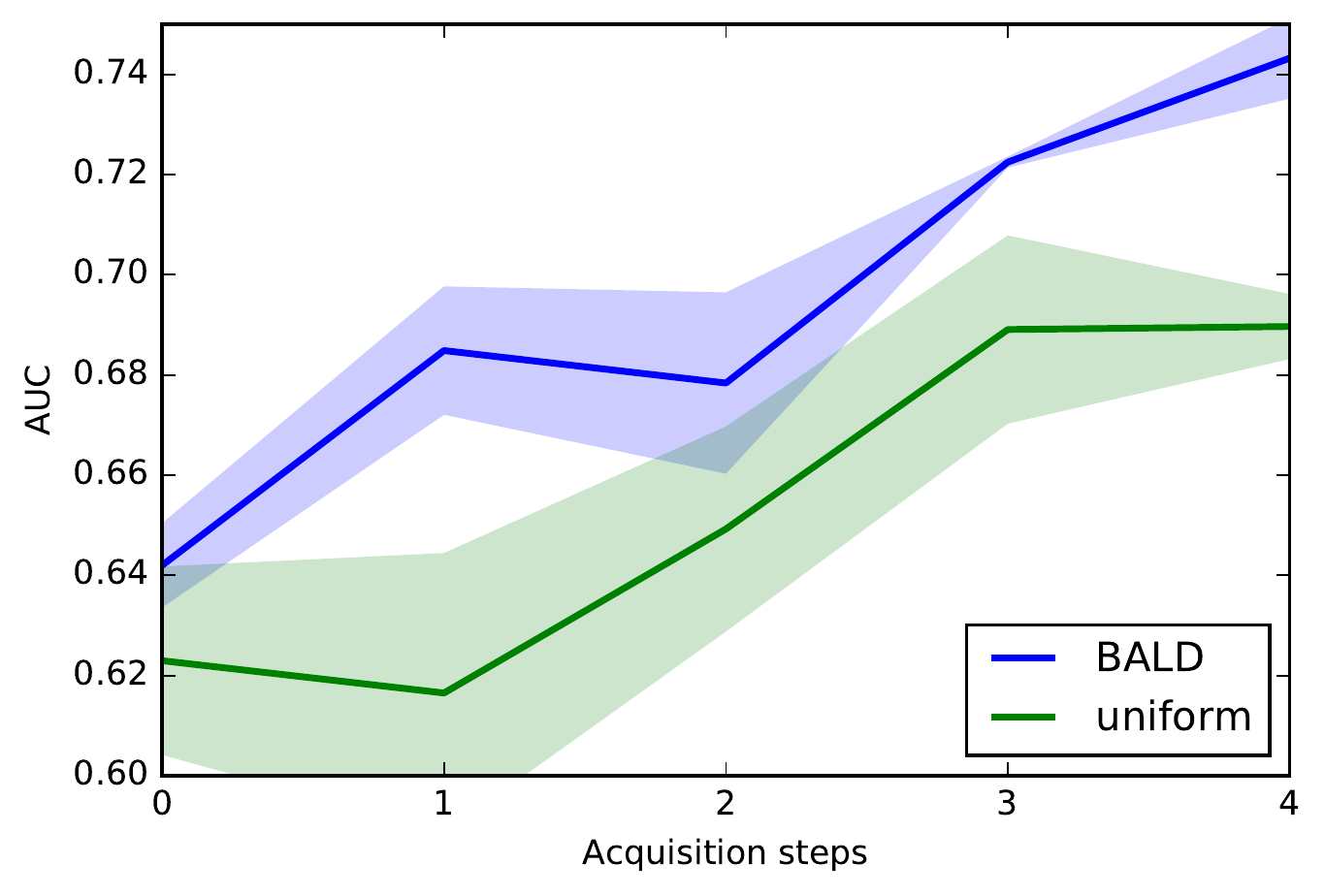}

\vspace{-3mm}
\caption{AUC as a function of acquisition step, first test split}
\end{subfigure}
\hspace{5mm}
\begin{subfigure}[t]{0.4\linewidth}
\includegraphics[width=\linewidth]{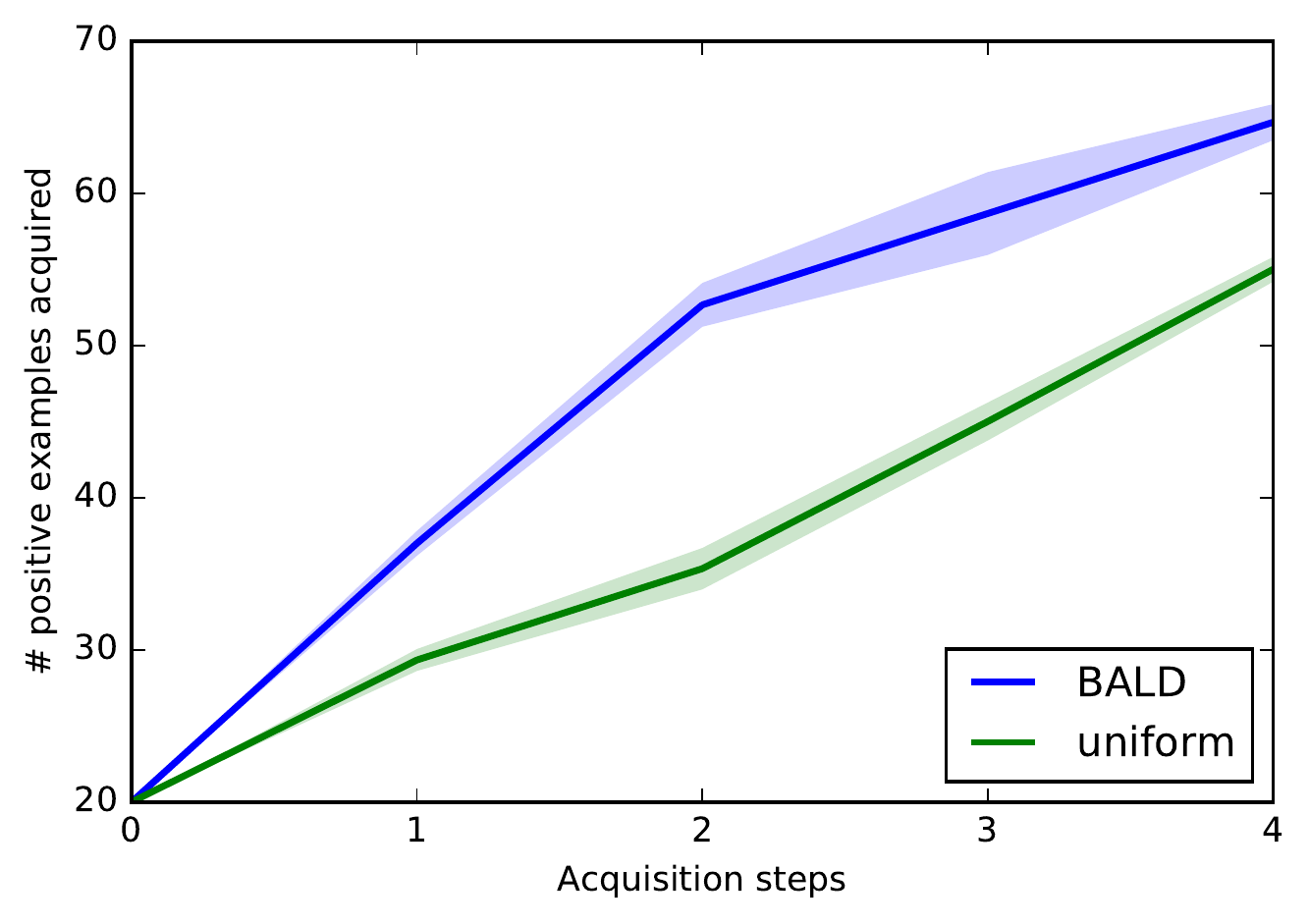}

\vspace{-3mm}
\caption{\# of positive examples as a function of acquisition step, first test split}
\end{subfigure}

\vspace{4mm}
\begin{subfigure}[t]{0.4\linewidth}
\includegraphics[width=\linewidth]{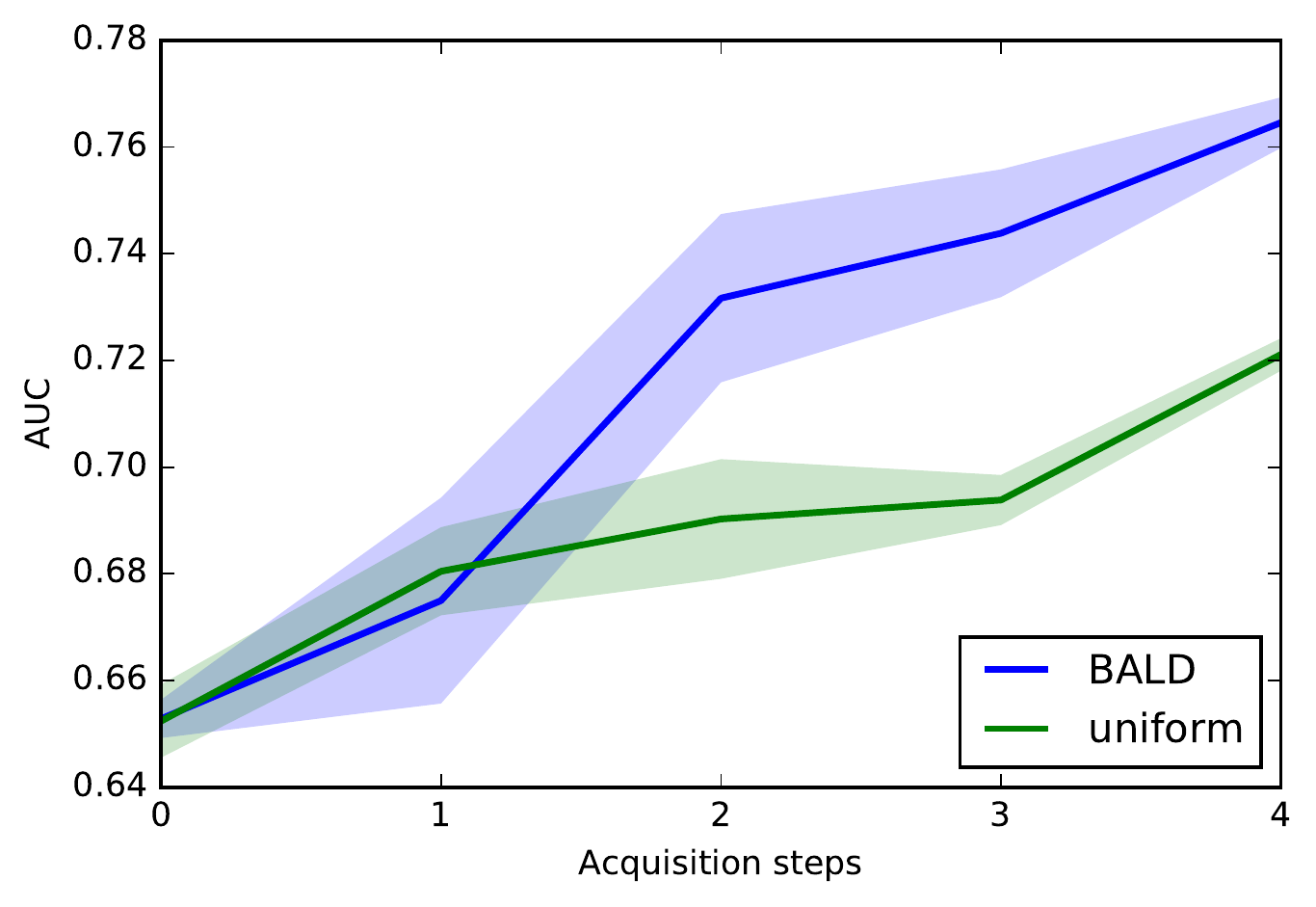}

\vspace{-3mm}
\caption{AUC as a function of acquisition step, second test split}
\end{subfigure}
\hspace{5mm}
\begin{subfigure}[t]{0.4\linewidth}
\includegraphics[width=\linewidth]{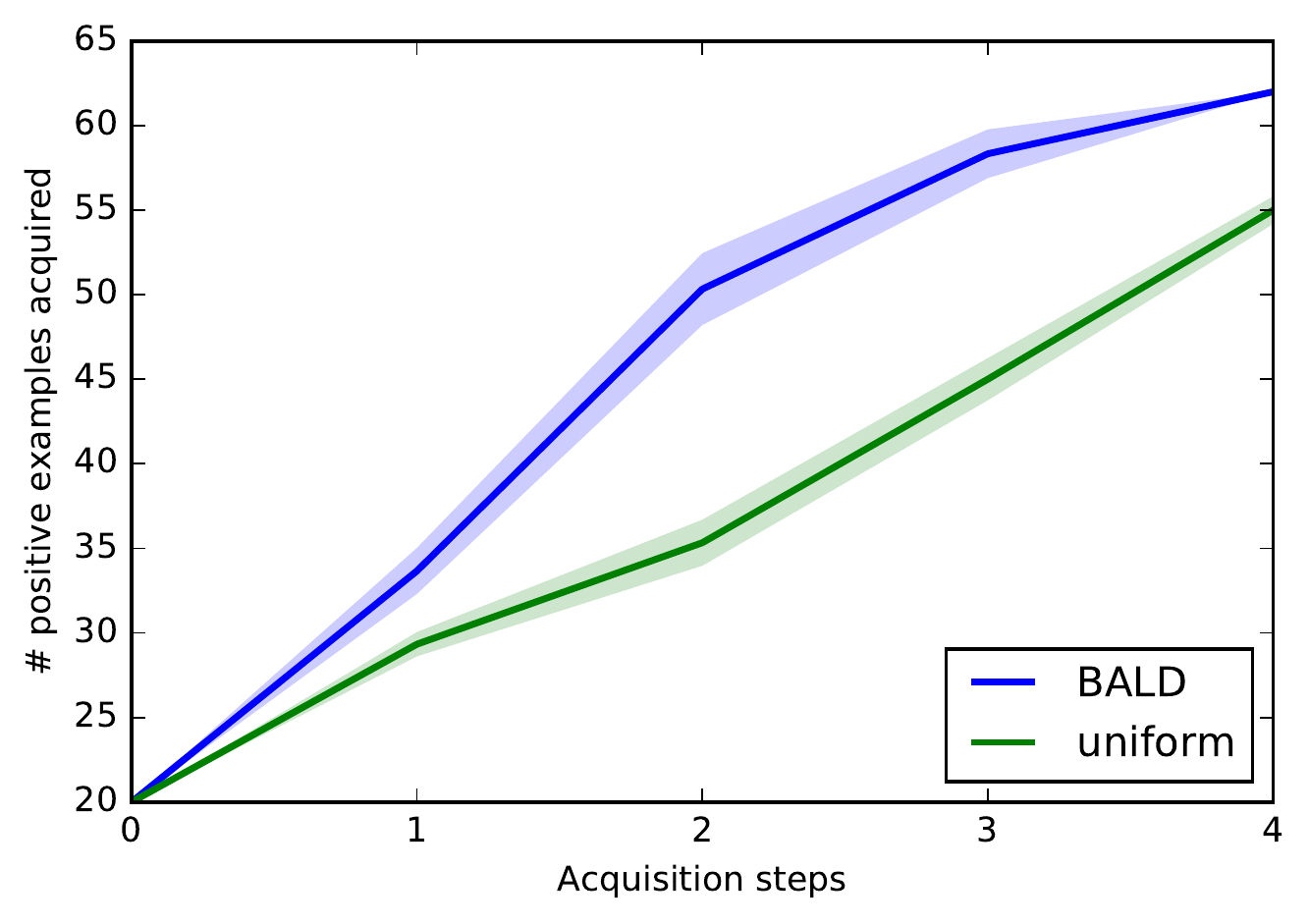}

\vspace{-3mm}
\caption{\# of positive examples as a function of acquisition step, second test split}
\end{subfigure}

\vspace{2mm}
\caption{AUC (left) as well as the number of acquired positive examples (right) for both the BALD acquisition function as well as uniform acquisition function, on \textbf{ISIC 2016 melanoma diagnosis dataset}. Two random test splits are assessed (top and bottom), and on each test set the experiment was repeated three times with different random seeds (shown mean with standard error).}
\label{fig:melanoma_res}
\end{figure*}

We experiment with active learning by following the following procedure. We begin by creating an initial training set of 80 negative examples and 20 positive examples from our training data, as well as a pool set from the remaining data. With each experiment repetition (out of the three experiment repetitions w.r.t.\ the fixed test split) the pool is shuffled anew. 
The positive examples in the current training set are augmented following the original training procedure, and a model is trained on the augmented training set for 100 epochs until convergence. We use batch size 8 and weight decay set by $(1 - p) l^2 / N$, where $N$ is the number of training points, $p=0.5$ is the dropout probability, and the length-scale squared $l^2$ is set to $0.5$. 
An acquisition function is then used to select the 100 most informative images from the pool set. These points are removed from the pool set and added to the (non-augmented) training set, where we use the original expert-provided labels for these points. The process is repeated until all pool points have been exhausted, where at each acquisition step we reset the model to its original pre-trained weights (as we also did in the previous section experiments). This reset is done in order to avoid local optima, and to avoid confusing model performance improvement with an improvement resulting from simply using longer (cumulative) optimisation time.

After each acquisition the test performance of the model is logged using MC dropout with 20 samples. We further keep track of the number of positive examples acquired after each acquisition. 
Model performance is assessed using area-under-the-curve (AUC) as this seems to be the most informative of all metrics used by \citet{gutman2016skin}. We experimented with the \textit{average precision} metric suggested by \citet{gutman2016skin} as well, but managed to get results improving over the competition winner by simply predicting all points as ``benign''. This might be because of the data imbalance.
AUC on the other hand takes into account all possible decision-thresholds possible to classify a malignant image. 

We assessed two acquisition functions: a uniform baseline, and BALD. Even though Variation Ratios performs well on MNIST above, the function fails with the melanoma data since most malignant images are given only a slight higher probability of being malignant compared to the probability of benign images of being malignant. As a result all pool points are given identical Variation Ratios acquisition value.

Experiment results are given in fig.\ \ref{fig:melanoma_res}, where results are reported on both test splits (top and bottom), and where with each split the experiment is repeated three times and performance results are averaged on that fixed split. 
For each test split we report mean with standard error.
AUC is reported for each split (left), and number of acquired positive examples is reported as well (right) for each acquisition step. 
BALD achieves better AUC faster than uniform, and acquires more positive examples at each acquisition step than uniform (i.e.\ BALD finds positive examples as informative and adds these to the training set, whereas uniform simply selects positive examples from the pool set based on their frequency). 

Note how AUC range varies wildly between the two different test splits, but how AUC is similar for both acquisition functions on each fixed test set before the initial acquisition (when both uniform and BALD models are trained on the same initial training set). This demonstrates the difficulties with handling of small data: each test split gives radically different results, and in this case even though each acquisition function experiment has a relatively small standard error, averaging the AUC of the acquisition functions over the different test splits would artificially increase the standard error. 
Lastly, it is interesting to experiment with a model trained over the entire pool set, i.e.\ with the settings of the second place winner in the ISIC2016 task.
For the first test split this model attains AUC $0.71 \pm 0.003$, whereas with the second test split it attains AUC $0.75 \pm 0.01$. For both test splits this AUC is worse than BALD's converged AUC after 4 acquisition steps. This might be because BALD avoided selecting noisy points -- near-by images for which there exist multiple noisy labels of different classes. Such points have large aleatoric uncertainty -- uncertainty which cannot be explained away -- rather than large epistemic uncertainty -- the uncertainty which BALD captures \textit{in order} to explain it away, i.e.\ reduce it.

\section{Future Research}

We presented a new approach for active learning of image data, relying on recent advances at the intersection of Bayesian modelling and deep learning, and demonstrated a real-world application in medical diagnosis. 
We assessed the performance of the techniques by resetting the models after each acquisition, and training them again to convergence. This was done to isolate the effects of our acquisition functions, which came at a cost of prolonged training times (20 hours for each melanoma experiment for example). We showed that even with this long running time, our technique still reduces required expert labels, thus reduces costs for such a system. This running time can be reduced further by not resetting the system -- with the potential price of falling into local optima. We leave this problem for future research.

\bibliography{references}
\bibliographystyle{icml2017}

\end{document}